\documentclass[10pt,twocolumn,letterpaper]{article}

\usepackage[pagenumbers]{cvpr} 

\definecolor{cvprblue}{rgb}{0.21,0.49,0.74}
\usepackage[pagebackref,breaklinks,colorlinks,allcolors=cvprblue]{hyperref}

\PassOptionsToPackage{table, dvipsnames}{xcolor}
\usepackage{multirow}
\usepackage[table]{xcolor}
\usepackage{colortbl}
\usepackage{graphicx}  
\pdfoutput=1

\newcommand\blfootnote[1]{%
  \begingroup
  \renewcommand\thefootnote{}\footnote{#1}%
  \addtocounter{footnote}{-1}%
  \endgroup
}

\begin{document}
\title{RAWMamba: Unified sRGB-to-RAW De-rendering With State Space Model}

\author{
Hongjun Chen$^{*}$,
Wencheng Han$^{*}$,
Huan Zheng,
Jianbing Shen$^{\dagger}$\\
SKL-IOTSC, CIS, University of Macau \\
{\tt\small hj.chen96@outlook.com}, {\tt\small \{wencheng256, huanzheng1998\}@gmail.com}, {\tt\small jianbingshen@um.edu.mo}
}
\maketitle

\blfootnote{$*$Equal contribution. $\dagger$Corresponding author: \textit{Jianbing Shen}.}   

\vspace{-4mm}
\begin{abstract}
Recent advancements in sRGB-to-RAW de-rendering have increasingly emphasized metadata-driven approaches to reconstruct RAW data from sRGB images, supplemented by partial RAW information. In image-based de-rendering, metadata is commonly obtained through sampling, whereas in video tasks, it is typically derived from the initial frame. The distinct metadata requirements necessitate specialized network architectures, leading to architectural incompatibilities that increase deployment complexity. In this paper, we propose {RAWMamba}, a Mamba-based unified framework developed for sRGB-to-RAW de-rendering across both image and video domains. The core of RAWMamba is the Unified Metadata Embedding (UME) module, which harmonizes diverse metadata types into a unified representation. In detail, a multi-perspective affinity modeling method is proposed to promote the extraction of reference information. In addition, we introduce the Local Tone-Aware Mamba (LTA-Mamba) module, which captures long-range dependencies to enable effective global propagation of metadata. Experimental results demonstrate that the proposed RAWMamba achieves state-of-the-art performance, yielding high-quality RAW data reconstruction.
\end{abstract}
    
\vspace{-2mm}
\section{Introduction}
\label{sec:intro}
RAW data directly captured from unprocessed camera sensor outputs, typically retains a higher bit depth (e.g., 14–16 bits) compared to 8-bit sRGB images, offering a significantly wider dynamic range that preserves more details across diverse lighting conditions~\cite{liu2020single}. 
Moreover, RAW data maintains a linear mapping of the original scene information, unlike the non-linear transformations introduced by Image Signal Processing (ISP) in sRGB images.
Recent studies have leveraged RAW data for tasks such as denoising and demosaicing~\cite{denoise1, denoise2, denoise3}, image super-resolution~\cite{xu2020exploiting, jiang2024rbsformer}, image classification~\cite{maxwell2024logarithmic},
image segmentation~\cite{cui2025raw}, and video super-resolution~\cite{jeelani2023expanding}, consistently achieving superior performance over sRGB-based approaches.
These findings underscore the substantial potential of RAW data to advance various computer vision tasks~\cite{cui2025raw, jeelani2023expanding}. However, RAW data demands considerably more memory and bandwidth than compressed formats like JPEG-based sRGB images, limiting its broader adoption. 
Consequently, more researchers~\cite{highquality, cam} focused on developing sRGB-to-RAW de-rendering algorithms to reconstruct RAW data from sRGB images.

\begin{figure}[t]
    \centering
    \includegraphics[width=0.49\textwidth]{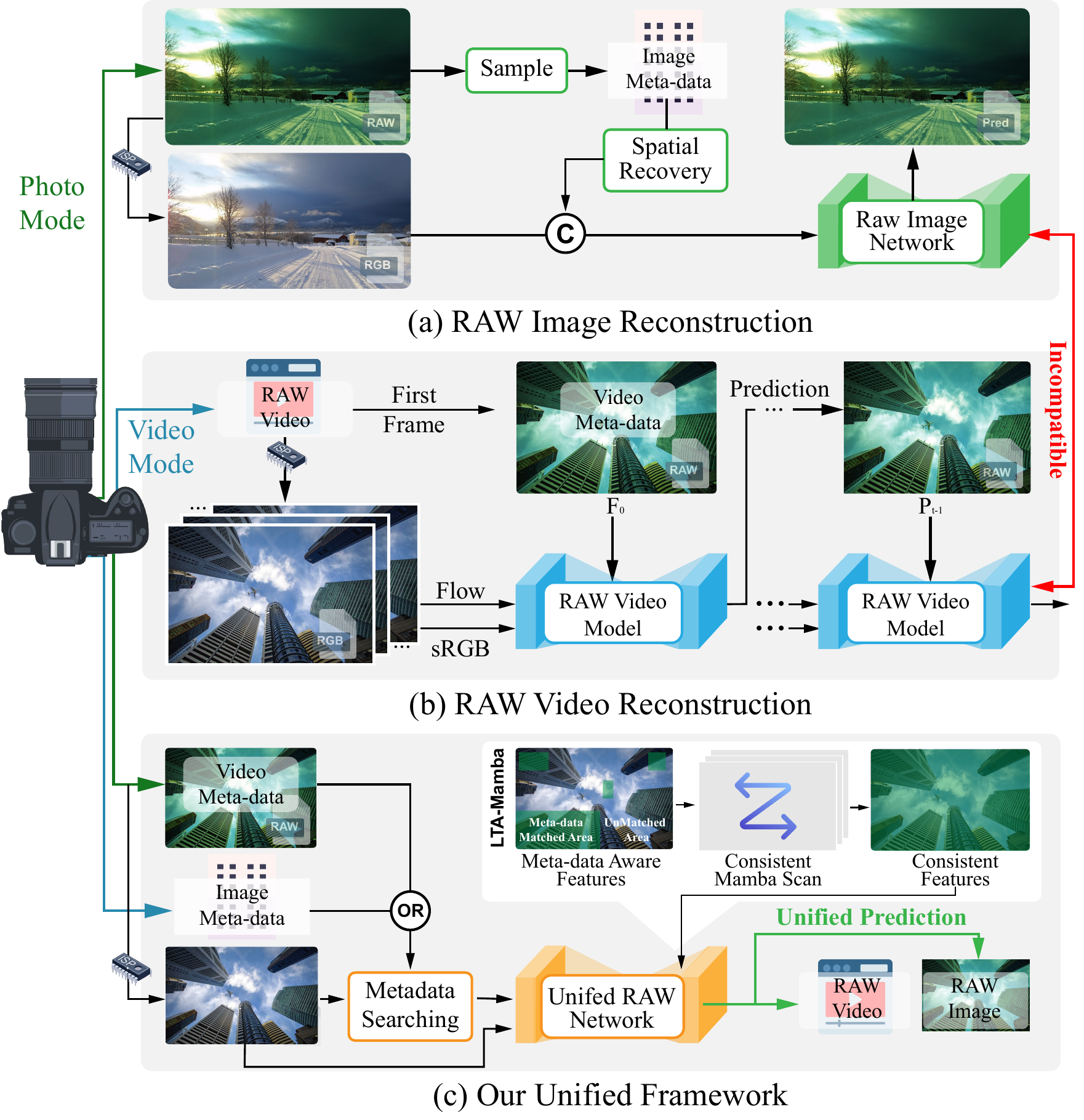} 
     \vspace{-5mm}
    \caption{\textbf{Comparison with Previous sRGB-to-RAW De-rendering Approaches.} (a) In image de-rendering methods like~\cite{spatiallyaware, cam, li2023metadata}, sampled RAW data is utilized as metadata, followed by spatial recovery operations to extract information. (b) For video de-rendering methods~\cite{videoraw}, the first frame serves as metadata, with a sequential model applied for information extraction. (c) Our RAWMamba method presents a unified framework capable of handling both image and video inputs.} 
    \vspace{-5mm}
    \label{fig:intro}  
\end{figure}

Some blind image sRGB-to-RAW de-rendering algorithms~\cite{debevec2008recovering, grossberg2003determining, liu2020single, xing2021invertible} rely solely on sRGB images to reconstruct RAW images. 
As the ISP process is inherently lossy, these methods face intrinsic limitations, often resulting in inferior reconstruction quality.
In contrast, more reliable approaches have been proposed to extract a subset of RAW data as metadata, serving as a reference to guide the reconstruction process.
Earlier methods~\cite{yuan2011high} utilized low-resolution RAW images or ISP parameters~\cite{nguyen2016raw} as metadata. 
More recent approaches~\cite{li2023metadata, spatiallyaware, cam} achieved metadata extraction by selectively sampling RAW data, aiming to retain essential information while minimizing memory requirements.
Such metadata are generally unsuitable for video with real-time, long-term requirements, as their per-frame processing introduces additional computational overhead on the camera. 
To address this, methods like~\cite{videoraw} use the first frame of an sRGB-RAW pair as metadata, improving both performance and efficiency compared to image-specific approaches.

To accommodate different input modalities and metadata, task-specific architectures are typically employed for sRGB-to-RAW de-rendering. 
As shown in \cref{fig:intro} (a) and (b), image-based tasks generally use spatial recovery methods to process sampling-based metadata, whereas video-based de-rendering solutions rely on a chained network architecture to propagate metadata from the first frame across subsequent frames. 
However, these image- and video-based models are incompatible with one another. 
Since most modern digital cameras can capture both video and image data, this lack of compatibility in de-rendering models makes the application less feasible.
Thus, \textit{a unified model capable of handling both video and image data is crucial for practical de-rendering applications}.

To achieve this, we propose RAWMamba,  the first Mamba-based unified RAW derendering model across both image and video domains. As shown in \cref{fig:intro} (c), our approach introduces a Unified Metadata Embedding (UME) module, that learns to extract reference information from different metadata sources. And a Local Tone-Aware Mamba (LTA-Mamba) network to enhance global consistency of the reference information in the reconstruction procedure.
In the UME module, we treat all types of metadata as a unified {(sRGB, RAW)} pair and establish a mapping between the de-rendering sRGB and the sRGB component within the metadata. 
Since sRGB and RAW images represent two modalities of the same data, this mapping is inherently applicable to the RAW space as well. 
Based on this, we leverage the RAW component of the metadata to extract aligned reference information for the de-rendering process. 
In detail, the UME module consists of two branches: one branch utilizes an affinity matrix to capture global metadata information, and another one employs dynamic position encoding with a deformable transformer block to extract local metadata information. 
This design aims to comprehensively extract reconstruction cues from the ISP pipeline, incorporating both global operations (e.g., white balance, gamma correction, and global denoising) and local operations (e.g., local tone mapping and local detail enhancement).

Our LTA-Mamba module is built upon the recent popular Mamba models~\cite{gu2023mamba,videomamba,zhu2024vision}, which have drawn significant attention in vision tasks for their abilities to efficiently model long-range dependencies.
This capability allows our model to enhance global consistency by leveraging sparse metadata references. 
Unlike the conventional use of Mamba~\cite{zhu2024vision, liu2024vmamba} in visual tasks, LTA-Mamba is specifically designed to address the challenge of local tone mapping in sRGB-to-RAW de-rendering. 
It integrates a local spatio-temporal Mamba block, a redundant feature aggregation mechanism, and a global spatio-temporal Mamba block.
Experiments demonstrate that our method achieves superior performance for sRGB-to-RAW de-rendering tasks. Specifically, when evaluated on the RVD-Part2~\cite{videoraw} video dataset and the CAM~\cite{cam} image dataset, RAWMamba achieves PSNR improvements of 2.20dB and 3.37dB, respectively, outperforming the task-specific models in both domains.
Overall, our contributions are listed as follows:
\begin{itemize}
    \item We are the first to introduce the Mamba structure into sRGB-to-RAW de-rendering tasks. We propose RAWMamba, a unified sRGB-to-RAW de-rendering network that effectively reconstructs both RAW images and videos using corresponding metadata.
    \item We propose the UME module to enable unified extraction of reference information from different types of metadata.
    \item We develop a Mamba-based module, LTA-Mamba, to effectively address the sparsity issues in metadata. 
    \item Experimental results demonstrate that RAWMamba achieves state-of-the-art performance on both video and image datasets, surpassing previous task-specific models and delivering high-quality RAW data reconstruction.
    
\end{itemize}

\section{Related Work}
RAW data typically exhibit a linear relationship with scene irradiance, offering advantages over sRGB images. This characteristic is widely leveraged in low-level vision tasks~\cite{brooks2019unprocessing, wei2020physics, xing2021end} and visual understanding applications~\cite{yu2021reconfigisp, xu2023toward, chen2023instance}. However, the large storage requirements and challenges in obtaining RAW data have led to increased interest in generating data images from sRGB images through inverse processing in recent years.

\noindent\textbf{Image sRGB-to-RAW de-rendering.}
Blind de-rendering refers to methods that rely solely on sRGB images for sRGB-to-RAW de-rendering. Early approaches~\cite{debevec2008recovering, grossberg2003determining} employed radiometric calibration to recover response functions from multiple exposure images. Subsequently, more complex methods~\cite{brooks2019unprocessing, conde2022model, kim2024paramisp} constructed reverse ISP workflows to reconstruct RAW images. Many methods leveraged the modeling power of deep learning. For instance, works in~\cite{liu2020single, xing2021invertible} proposed an invertible ISP network to map between sRGB and RAW image spaces.
However, the information loss during the ISP process unavoidably limits the performance of these reconstruction methods.

To further improve the reliability of sRGB-to-RAW de-rendering, recent studies have explored leveraging a small amount of metadata to assist on reconstruction. 
Nguyen and Brown~\cite{nguyen2016raw, nguyen2018raw} proposed a straightforward idea to store the ISP pipeline parameters as metadata. However, their work overlooks some ISP operations that cannot be quantitatively stored, such as local tone mapping. Consequently, the mainstream way to obtain metadata is sampling-based methods.
Early sampling method~\cite{yuan2011high} storage low-resolution RAW images as metadata for reducing cost, and up-sampled the lower-resolution RAW images during the reconstruction phase.
Punnappurath and Brown~\cite{spatiallyaware} reduced the size of metadata by treating a small set of uniformly sampled raw pixels as metadata. 
Subsequently, Li \textit{et al.}~\cite{li2023metadata} further used implicit neural functions to up-sampling the uniformly sampled raw pixels and make a significant improvement.
\cite{cam} further optimized the sampling strategy by selecting representative raw pixels based on superpixels. It jointly trained the sampler and reconstructor in an end-to-end framework, achieving notable performance at a sparse sampling rate.
Recently, Wang \textit{et al.}~\cite{wang2024beyond} proposed an adaptive and learnable metadata construction method, and utilized a compact representation of the latent space in place of sampled pixels.
For continuously generated videos, using sampling strategies to obtain metadata meets two primary challenges: i) the extensive, high-detail information embedded in RAW data requires considerable storage capacity, and ii) capturing RAW video data in real-time imposes a substantial burden on the camera's bandwidth.
Zhang \textit{et al.}~\cite{videoraw} further proposed using a single RAW frame from the video, typically the first frame of sRGB-RAW pair as metadata. 
To utilize such metadata, they employ a chained model, where each frame utilizes the predicted RAW data from the previous frame to generate its de-rendered RAW data.

\noindent\textbf{Mamba models for vision tasks.}
Recently, State Space Models (SSMs)~\cite{kalman1960} have regained attention in artificial intelligence research, owing to their remarkable efficiency in using state space transformations~\cite{gu2021combining} to handle long-term dependencies in language sequences. 
To relieve the cost burden of SSMs, S4~\cite{gu2021efficiently} introduced a structured state-space sequence to efficiently compute long-term dependencies. Mamba~\cite{gu2023mamba} and Mamba2~\cite{mamba2} further introduces a hardware-aware parallel algorithm (S6), offering the benefits of fast inference and linear scaling with sequence length. It significantly outperforms transformer~\cite{transformer} architectures in terms of model parameter efficiency.
Recently, Mamba has been adopted for constructing vision foundation models~\cite{zhu2024vision, liu2024vmamba, guo2025mambair} and introduces various scanning mechanisms tailored to the characteristics of 2D images, including bi-directional scan~\cite{zhu2024vision}, cross-scan~\cite{liu2024vmamba}, local scan~\cite{huang2024localmamba, rainmamba, Xu2024LGRNetLR}.
Due to the exceptional ability to tackle complex vision challenges, SSMs have been integrated into vision understanding tasks~\cite{huang2024localmamba, xing2024segmamba, mtmamba} and low level vision tasks~\cite{guo2025mambair, shi2024vmambair, rainmamba}.
With the ability to retain long-term dependencies, SSMs are well-suited for handling temporal relationships, making them a natural choice for tasks like video object segmentation~\cite{yang2024vivim, Xu2024LGRNetLR}, video understanding~\cite{videomamba, videomambapro} and motion generation~\cite{motionmamba}.

To the best of our knowledge, SSMs have not yet been explored in the sRGB-to-RAW de-rendering task. 
Distinct from previous SSM-based works, we introduce a local tone mapping-aware design to SSMs, specifically improving RAW reconstruction quality.

\section{Method}
\begin{figure*}[htbp]
    \centering
    \includegraphics[width=0.98\linewidth]{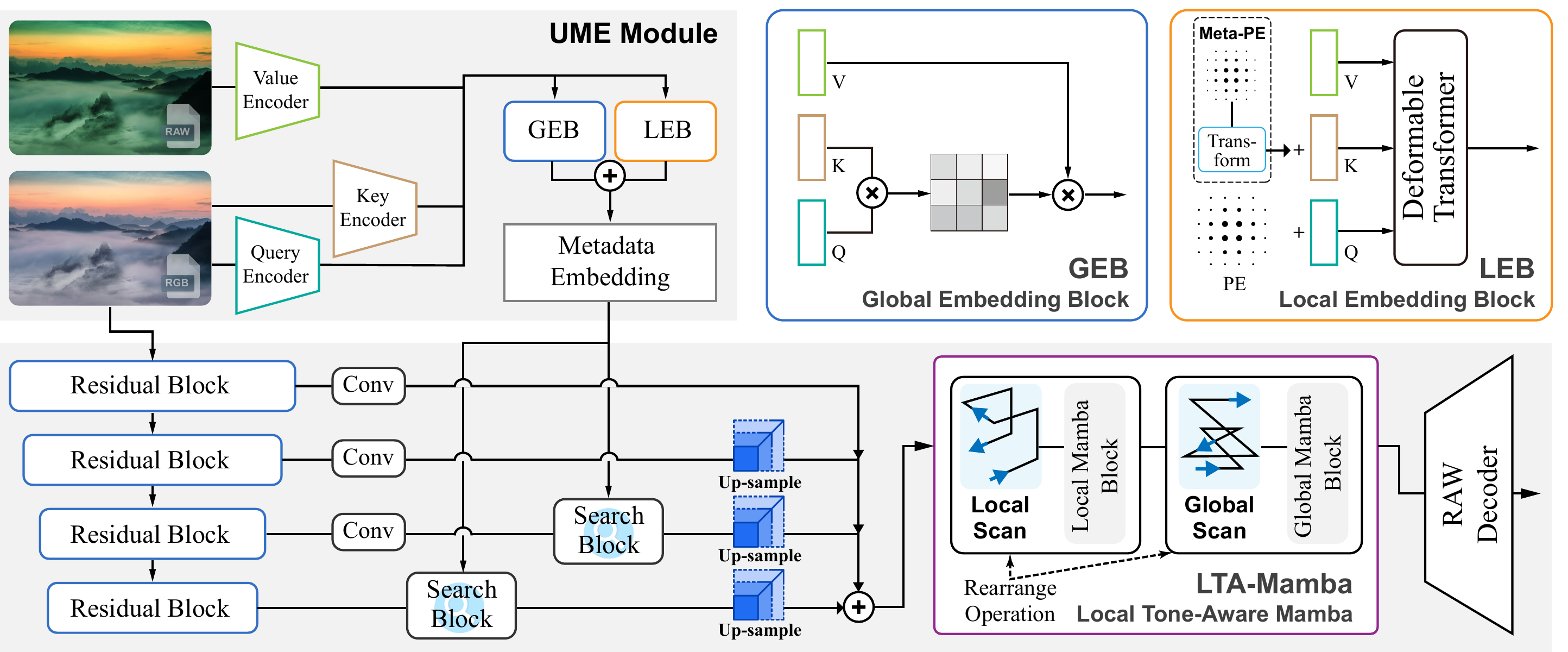} 
    \caption{\textbf{Overview of the Proposed RAWMamba.} In RAWMamba, inputs are processed through the UME module and the main network. The UME module encodes images and metadata into feature spaces, generating global and local metadata embeddings via affinity-based blocks (GEB and LEB). The main network query the metadata embeddings to refine features, which are aggregated input into the LTA-Mamba module. LTA-Mamba employs local and global spatiotemporal scanning to enhance global consistency for the final reconstruction. }
    \label{fig:pipeline}  
    \vspace{-3mm}
\end{figure*}

In this section, we provide an overview of the proposed RAWMamba framework, as show in \cref{fig:pipeline}. Specifically, we first offer a brief introduction to Mamba, and then proceed to describe the two key components of RAWMamba including UME and LTA-Mamba.

\subsection{Preliminaries}
SSMs are inspired by continuous linear time-invariant systems and are designed to efficiently map one-dimensional sequences $x(t) \in \mathbb{R} \mapsto y(t) \in \mathbb{R}$ via a hidden state $h(t) \in \mathbb{R}^\mathtt{N}$, where $\mathtt{N}$ is the size of hidden state. SSMs model the input data by employing the following Ordinary Differential Equation (ODE):
\begin{equation}
    \begin{aligned}
        \label{eq:lti}
        h'(t) &= \mathbf{A}h(t) + \mathbf{B}x(t), \\
        y(t) &= \mathbf{C}h(t),
    \end{aligned}
\end{equation}
where $\mathbf{A} \in \mathbb{R}^{\mathtt{N} \times \mathtt{N}}$ denotes the evolution parameter and $\mathbf{B} \in \mathbb{R}^{\mathtt{N} \times 1}$, $\mathbf{C} \in \mathbb{R}^{1 \times \mathtt{N}}$ denotes the projection parameters.

Mamba~\cite{gu2023mamba} is one of the discrete versions of the presented continuous system, which includes a timescale parameter $\mathbf{\Delta}$ to transform the continuous parameters $\mathbf{A}$, $\mathbf{B}$ to their discrete representation $\mathbf{\overline{A}}$, $\mathbf{\overline{B}}$. Mamba used Zero-Order Hold (ZOH) method for transformation, defined as:
\begin{equation}
    \begin{aligned}
        \label{eq:zoh}
        \mathbf{\overline{A}} &= \exp{(\mathbf{\Delta}\mathbf{A})}, \\
        \mathbf{\overline{B}} &= (\mathbf{\Delta} \mathbf{A})^{-1}(\exp{(\mathbf{\Delta} \mathbf{A})} - \mathbf{I}) \cdot \mathbf{\Delta} \mathbf{B}.
    \end{aligned}
\end{equation}
Subsequently, using a step size of $\mathbf{\Delta}$, we can discretize Eq.~\eqref{eq:lti} as follows:
\begin{equation}
    \begin{aligned}
        \label{eq:discrete_lti}
        h_t &= \mathbf{\overline{A}}h_{t-1} + \mathbf{\overline{B}}x_{t}, \quad y_t = \mathbf{C}h_t.
    \end{aligned}
\end{equation}
Finally, the iterative process outlined in Eq.~\eqref{eq:discrete_lti} can be efficiently computed in parallel through a global convolution:
\begin{equation}
    \begin{aligned}
        \label{eq:conv}
        \mathbf{y} &= \mathbf{x} \circledast \mathbf{\overline{K}}, \\
        with \quad \mathbf{\overline{K}} &= (\mathbf{C}\mathbf{\overline{B}}, \mathbf{C}\mathbf{\overline{A}}\mathbf{\overline{B}}, \dots, \mathbf{C}\mathbf{\overline{A}}^{\mathtt{L}-1}\mathbf{\overline{B}}),  
    \end{aligned}
\end{equation}
where $\mathtt{L}$ is the length of the input sequence $\mathbf{x}$, $\circledast$ represents the convolution operation, and $\overline{\mathbf{K}} \in \mathbb{R}^{\mathtt{L}}$ is a convolutional kernel. Such approach enables Mamba to handle long-term sequences more efficiently.

\subsection{sRGB-to-RAW De-rendering}
Metadata-based sRGB-to-RAW de-rendering aims to reconstruct RAW video sequences or images from sRGB data and limited metadata. The metadata provides a sparse representation of information from the original RAW data. Given a sRGB video sequence comprising $n$ frames, denoted as $x = \{x_1, x_2, \dots, x_n\}$, and its corresponding RAW video sequence $y = \{y_1, y_2, \dots, y_n\}$, the de-rendering process is modeled by the function $f$. The sRGB-to-RAW de-rendering procedure is defined as:
\begin{equation}
    y = f(x \mid \epsilon),
\end{equation}
where $\epsilon$ indicates the metadata, and we set the first frame$\{x_1,y_1\}$ as $\epsilon$. For the image sRGB-to-RAW de-rendering task, $n$ is set to 1, with $x$ and $y$ representing a single image, while $\epsilon$ is defined as sampled RAW pixels $y_s$. 

In this work, our de-rendering pipeline, as shown in \cref{fig:pipeline}, consists of two sub-networks: metadata embedding and reconstruction. First, we feed the input data $(x,\epsilon)$ into the UME module. Within the UME module, images and metadata are separately encoded into feature spaces, generating global and local metadata embeddings via affinity-based embedding blocks (GEB and LEB). 
In the main network, we use the encoded features of $x$ to query the metadata embeddings, then aggregate multi-scale features and input them into the LTA-Mamba module. The LTA-Mamba module employs global and local spatiotemporal scanning on the features to enhance global consistency. Finally, the enhanced features are passed through a reconstruction decoder, composed of multiple 3D convolutional layers and upsampling layers, to generate the reconstructed RAW data.

\subsection{Unified Metadata Embedding (UME)}
Image de-rendering algorithms~\cite{highquality, spatiallyaware, cam} often employ spatial recovery techniques to reconstruct metadata and integrate them directly into a neural network for learning. However, the success of these methods relies on the premise that the metadata and the input sRGB image are spatially aligned. In contrast, video de-rendering algorithms~\cite{videoraw} cannot employ the same strategy, as they use the first frame as metadata. Consequently, a common approach is to implement a chained structure for processing the video sequence, utilizing generated RAW data and optical flow information to guide the generation of subsequent frames. This highlights that models for image and video de-rendering are task-specific and inherently incompatible.

To address this issue, we propose the UME module, which is designed to extract reconstructive reference information in a unified manner from different types of metadata, thereby effectively obtaining metadata embedding. 
The core function of the UME module is to align various forms of metadata with sRGB.
To this end, we treat all types of metadata as a unified {(sRGB, RAW)} pair and establish a mapping between sRGB data and the sRGB component within the metadata. 
Since sRGB and RAW data represent two different modalities of the same information, this mapping is inherently applicable to the RAW domain as well. 
As a result we can extract aligned metadata information.

In detail, both image and video inputs are unified denoted as $(x_{srgb}, x_{srgb}^{'}, x_{raw}^{'})$. Here, $x_{srgb}$ represents the sRGB images, $x_{srgb}^{'}$ contains the sRGB part of metadata, and $x_{raw}^{'}$ represents the raw part of metadata. Specifically, these symbols in the video de-rendering task are defined as: 
\begin{equation}
    x_{srgb}^{'}=x_1, \quad x_{raw}^{'}=y_1,
\end{equation}
and in the image de-rendering task are defined as:
\begin{equation}
    x_{srgb}^{'} = x \cdot mask, \quad x_{raw}^{'} = y_s,
\end{equation}
where $mask$ represents sampling position. However, due to the absence of a direct alignment relationship between $x_{srgb}$ and $x_{srgb}^{'}$ in the image domain, we employ different encoders to map them to feature space. We use three convolutional neural networks (CNNs) based encoder, denoted as $f_1$, $f_2$, and $f_3$, to encode the raw and sRGB images into a shared latent feature space:
\begin{equation}
    \begin{aligned}
        &\mathbf{F}_{srgb} = f_1(x_{srgb}),  \\
        &\mathbf{F}_{srgb}^{'} = f_2(x_{srgb}^{'}), \\
        &\mathbf{F}_{raw}^{'} = f_3(x_{raw}^{'}),
    \end{aligned}
\end{equation}
where $\mathbf{F}_{srgb}$, $\mathbf{F}_{srgb}^{'}$ and $\mathbf{F}_{raw}^{'}$ are correspond intent features. We then introduce two sub-branches: the Global Embedding Block (GEB) and the Local Embedding Block (LEB), for leveraging referred mapping relationship to produce metadata embedding.

\noindent
\textbf{GEB.} 
The GEB is designed to capture global operation wise information within the ISP pipeline. 
We multiply the feature map of $x_{srgb}$ and $x_{srgb}^{'}$ to generate an affinity matrix $\mathbf{A}$ to represent their alignment relationship:
\begin{equation}
    \mathbf{A}_1 = \text{Softmax} ((\mathbf{F}_{srgb})^T  \mathbf{F}_{srgb}^{'}),
\end{equation}
where $\text{Softmax}$ is applied in the reference dimension. We then multiply affinity matrix with $\mathbf{F}_{raw}^{'}$ to integrate aligned reference information in a global perspective, and produce global metadata embedding $\mathbf{E}_{global}$:
\begin{equation}
    \mathbf{E}_{global} = \mathbf{A}_1 \mathbf{F}_{raw}^{'}.
\end{equation}


\noindent
\textbf{LEB.} The LEB is designed to capture local operation-specific information within the ISP pipeline. 
To enhance its focus on localized regions, we introduce position encoding in this branch.
In detail, we directly add the same position embeddings $\text{PE}$ in image-level task, since the spatial positions of image metadata are aligned with the de-rendering sRGB. In contrast, due to motion, the spatial alignment of video metadata and de-rendered sRGB is not inherently consistent. To address this, we apply optical flow to dynamically transform the position embeddings of the metadata, denoted as $\bar{\text{PE}}$, ensuring alignment with the dynamic motion in video sequences. Our position encoding method may introduce errors, so we employ a Deformable Transformer~\cite{zhu2020deformable} block to correct the alignment, thereby achieving improved local metadata embeddings. The process can be formalized as follows:

\begin{equation}
    \mathbf{A}_2 = \text{Softmax} \left( \frac{(\mathbf{F}_{srgb} + \text{PE}) (\mathbf{F}_{srgb}^{'\top} + \bar{\text{PE}})}{\sqrt{d}} + \mathbf{B} \right),
\end{equation}
where $d$ is the dimensionality of the features, $\mathbf{B}$ is the deformable offset and $\mathbf{A}_2$ is local attention matrix. We then produce local metadata embedding $\mathbf{E}_{local}$:
\begin{equation}
    \mathbf{E}_{local} = \mathbf{A_2} (\mathbf{F}_{raw}^{'}+\bar{\text{PE}}).
\end{equation}

By performing a weighted sum of $\mathbf{E}_{local}$ and $\mathbf{E}_{global}$, we obtain the complete metadata embedding $\mathbf{E}$. The main network then extracts valuable reference information from $\mathbf{E}$ through a search block that operates across multiple scales. This search block is implemented using a channel-wise transformer block.

\subsection{Local Tone-Aware Mamba (LTA-Mamba)}
\begin{figure}[t]
    \centering
    \includegraphics[width=0.49\textwidth]{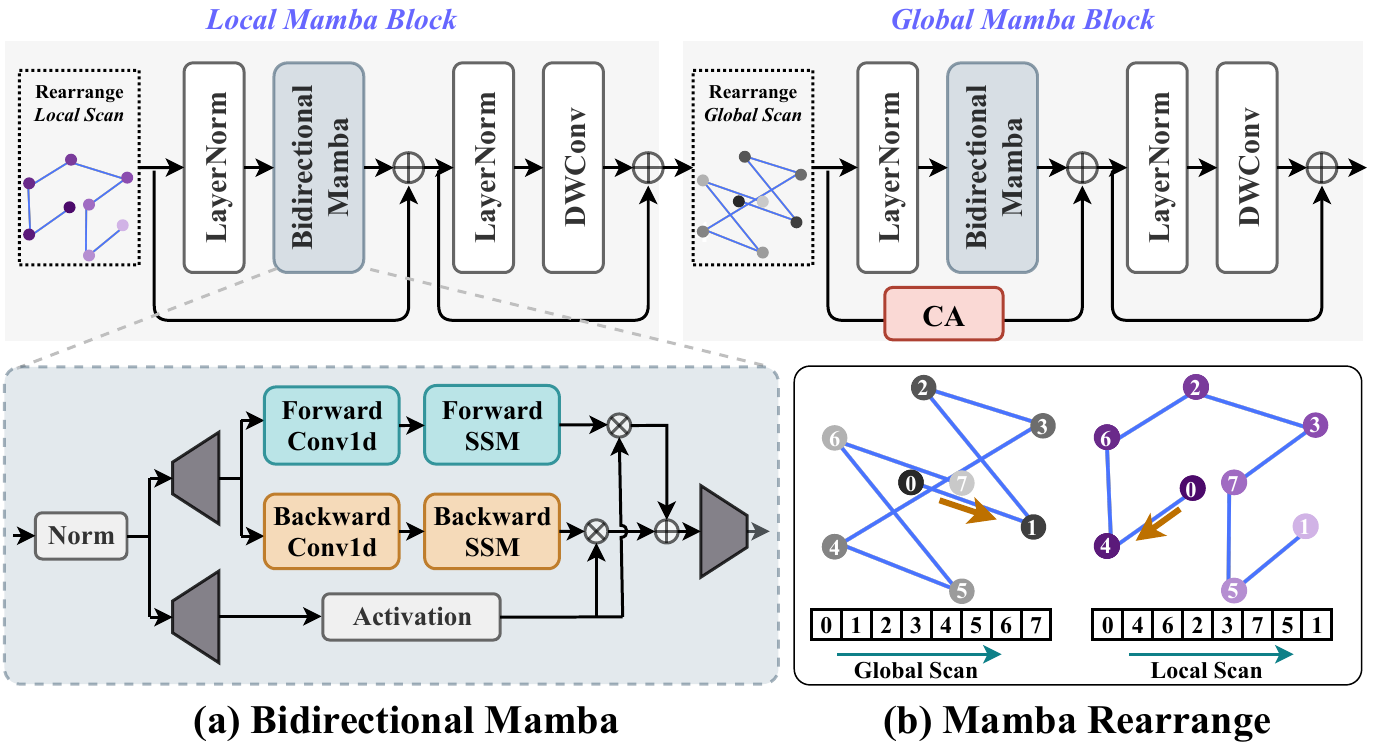} 
    \vspace{-4mm}
    \caption{\textbf{The details of the LTA-Mamba module.} The LTA-Mamba module relies on two consecutive Mamba blocks. In the figure, the DWConv block represents a depth-wise convolution layer, while CA denotes a channel attention block. (a) The implementation of bi-direction mamba. (b) An illustration of the global scan strategy and the local scan strategy.} 
    \label{fig:mamba}  
    \vspace{-9mm}
\end{figure}

The sparse nature of the metadata presents significant challenges in accurately reconstructing global RAW data from limited information. This sparsity necessitates modeling long-range dependencies to effectively propagate metadata information throughout the entire image or video. Previous CNN-based models~\cite{cam,videoraw} are constrained by their local spatiotemporal receptive fields, resulting in limited performance in capturing global context. Although transformers with multi-head attention mechanisms~\cite{transformer} enable a global understanding of sequences, their quadratic complexity with respect to sequence length poses efficiency challenges when processing pixel-level features.

To efficiently handle long-term dependencies across both image and video pixel sequences, we integrate the Mamba module into our network. Mamba offers linear complexity relative to sequence length and long-range modeling ability. By leveraging Mamba, we can propagate the influence of the sparse metadata across the entire dataset, effectively filling in gaps where metadata is absent. 

Specifically, we account for the local tone mapping characteristics of ISP and propose a Local Tone-Aware Mamba module (LTA-Mamba), shown in \cref{fig:mamba}, comprising a local-scan based spatiotemporal Mamba block, and a global-scan based spatiotemporal Mamba block.

\subsubsection{Local Spatiotemporal Mamba Block}
Given an input video of $n$ frames, we first aggregate the multi-level features $\{\mathbf{F}_{enc}^i \in \mathbb{R}^{(n \times C \times H/N_i \times W/N_i)} \}$ generated by the encoder, where $N_i$ denotes the downsampling factor, the aggregation process is as follows:
\begin{equation}
    \mathbf{F} = \text{Conv}(\sum_{i=1}^{k} up(\mathbf{F}_{enc}^i)),
\end{equation}
where $up_{\times N_i}$ denotes upsampling operation to align the size of feature maps, $k$ denotes number of aggregated output layers, and $\text{Conv}$ represents a convolutional layer for feature fusion. 

The aggregated feature map $\mathbf{F}$ is subsequently input into the first block of the LTA-Mamba module, referred to as the Local Spatiotemporal Mamba Block.

The primary challenge in sRGB-to-RAW de-rendering lies in the numerous hard-to-invert nonlinear operations inherent in the ISP, particularly local tone mapping~\cite{localtonemap}. The implementation of local tone mapping takes into account surrounding information of pixels, such as brightness and contrast, and applies adaptive mapping functions to different regions based on their local characteristics. To enhance the learning of its inversion process, we have aligned the design of our network architecture accordingly.

Initially, it is essential to capture the local dependencies of features to effectively mimic the local functions.
Models~\cite{zhu2024vision, liu2024vmamba} based on Mamba typically employ a line-by-line scanning strategy to construct sequences, which neglects local spatial relationships in 2D images and the spatiotemporal correlations inherent in video data.
In this paper, we employ a 3D Hilbert curve for localized scanning. The 3D Hilbert curve is recursively constructed, where each recursion divides the cube into eight sub-cubes, generating the curve within these sub-cubes in a specific order and direction. 
This approach enables us to obtain local perspective coordinates $\mathbf{C}_{\text{L}}$, which are then used to rearrange the sequence of feature $\mathbf{F}$:
\begin{equation}
    \Bar{\mathbf{F}} = \text{Rearrange}(\mathbf{F}, \mathbf{C}_{\text{L}}),
\end{equation}
where $\Bar{\mathbf{F}}$ is a sequence ordered in a local spatiotemporal continuous. Then $\Bar{\mathbf{F}}$ input our mamba block:
\begin{equation}
    \begin{aligned}
    \label{eq:ssm}
        \Bar{\mathbf{F}}_1 &= \text{SSM}(\text{LN}(\Bar{\mathbf{F}})) + \Bar{\mathbf{F}}, \\
        \Bar{\mathbf{F}}_2 &= \Bar{\mathbf{F}}_1 + \text{DWC}(\text{LN}(\Bar{\mathbf{F}}_1)),
    \end{aligned}    
\end{equation}
where $\text{LN}$ represents layer normalization operation, $\text{SSM}$ is a bidirectional state space model introduced by~\cite{zhu2024vision}, and $\text{DWC}$ is a depth-wise convolution layer to preserve fine-grained details.

\subsubsection{Global Spatiotemporal Mamba Block}
In the ISP pipeline, the accumulation of numerous local tone mapping functions constitutes a comprehensive global tone mapping. Similarly, after capturing local dependency information, global integration is necessary to achieve consistent output. 
We incorporate a global spatiotemporal mamba block that utilizes a horizontal line-by-line scanning approach. This global scanning strategy allows the Mamba to establish contextual dependencies between each pixel and every other pixel within a linearized sequence, ensuring robust global context consistency.

The construction of this block aligns with \cref{eq:ssm}. we just need to rearrange $\Bar{\mathbf{F}}_2$ with line-by-line scanning coordinates $\mathbf{C}_{\text{G}}$:
\begin{equation}
    \mathbf{F}_2 = \text{Rearrange}(\Bar{\mathbf{F}_2}, \mathbf{C}_{\text{G}}),
\end{equation}

It is noted that local mapping functions primarily focus on the characteristics within the neighborhood of corresponding region. 
While using neural networks for de-rendering, the latent layer features are often globally distributed, which introduces redundancy for restoring local tone mapping relationship. To enhance training efficiency, we add channel attention~\cite{senet} before learning long-term dependencies. 

In detail, the procedure in global spatiotemporal mamba block is defined as:
\begin{equation}
    \begin{aligned}
        \mathbf{F}_3 &= \text{SSM}(\text{LN}(\mathbf{F}_2)) + \mathbf{F}_2, \\
        \hat{\mathbf{F}}_2 &= \text{CA}(\mathbf{F}_2), \\
        \mathbf{F}_4 &= \mathbf{F}_3 + \text{DWC}(\text{LN}(\mathbf{F}_3 + \hat{\mathbf{F}}_2))),
    \end{aligned}
\end{equation}
where $\text{CA}$ is a channel attention operation. Finally, we refine $\mathbf{F}_4$ through a convolution layer. After iteratively applying multiple LTA-Mamba modules, the information within the metadata embedding propagate through the feature space, establishing global consistency. Then the consistency-aware feature maps are fed into the decoder for reconstruction.

For image-level inputs, we set $n=1$, allowing the model to learn global and local dependencies in the spatial dimension, thereby enhancing the reconstruction quality.

Our network is trained in a end-to-end manner. To demonstrate the effectiveness of its structure, we employed a commonly used loss function $\mathcal{L}$ that combines the Mean Squared Error (MSE) loss with the SSIM loss~\cite{ssim} to measure the visual quality, which is defined as:
\begin{equation}
    \mathcal{L} = \mathcal{L}_{mse} + \lambda \mathcal{L}_{ssim},
\end{equation}
where $\lambda$ is a hyperparameter used to balance the contribution of losses and is set to 0.5 in this paper.

\begin{figure*}[t]
	\centering
	\subfloat[Visual comparisons for \textbf{sRGB-to-RAW \textit{image} de-rendering} in Olympus E-PL6, Samsung NX2000, and Sony SLT-A57, respectively.]{%
        \begin{minipage}{\textwidth}
            \centering
            \footnotesize
            \includegraphics[width=0.18\textwidth,height=0.12\textwidth]{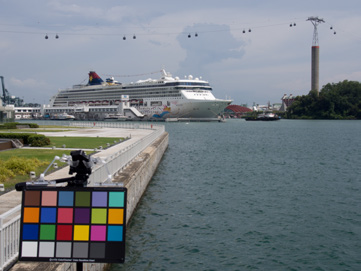}
            \includegraphics[width=0.18\textwidth,height=0.12\textwidth]{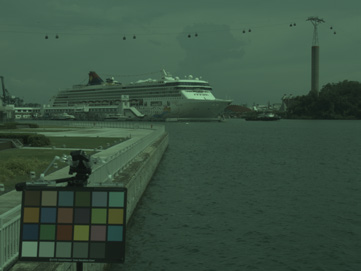}
            \includegraphics[width=0.18\textwidth,height=0.12\textwidth]{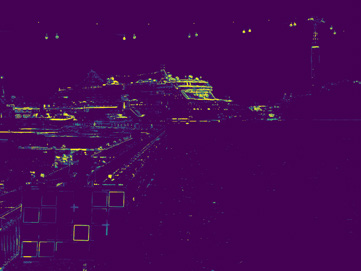}
            \includegraphics[width=0.18\textwidth,height=0.12\textwidth]{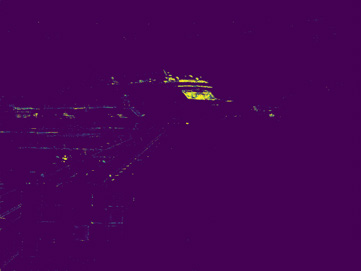}
            \includegraphics[width=0.18\textwidth,height=0.12\textwidth]{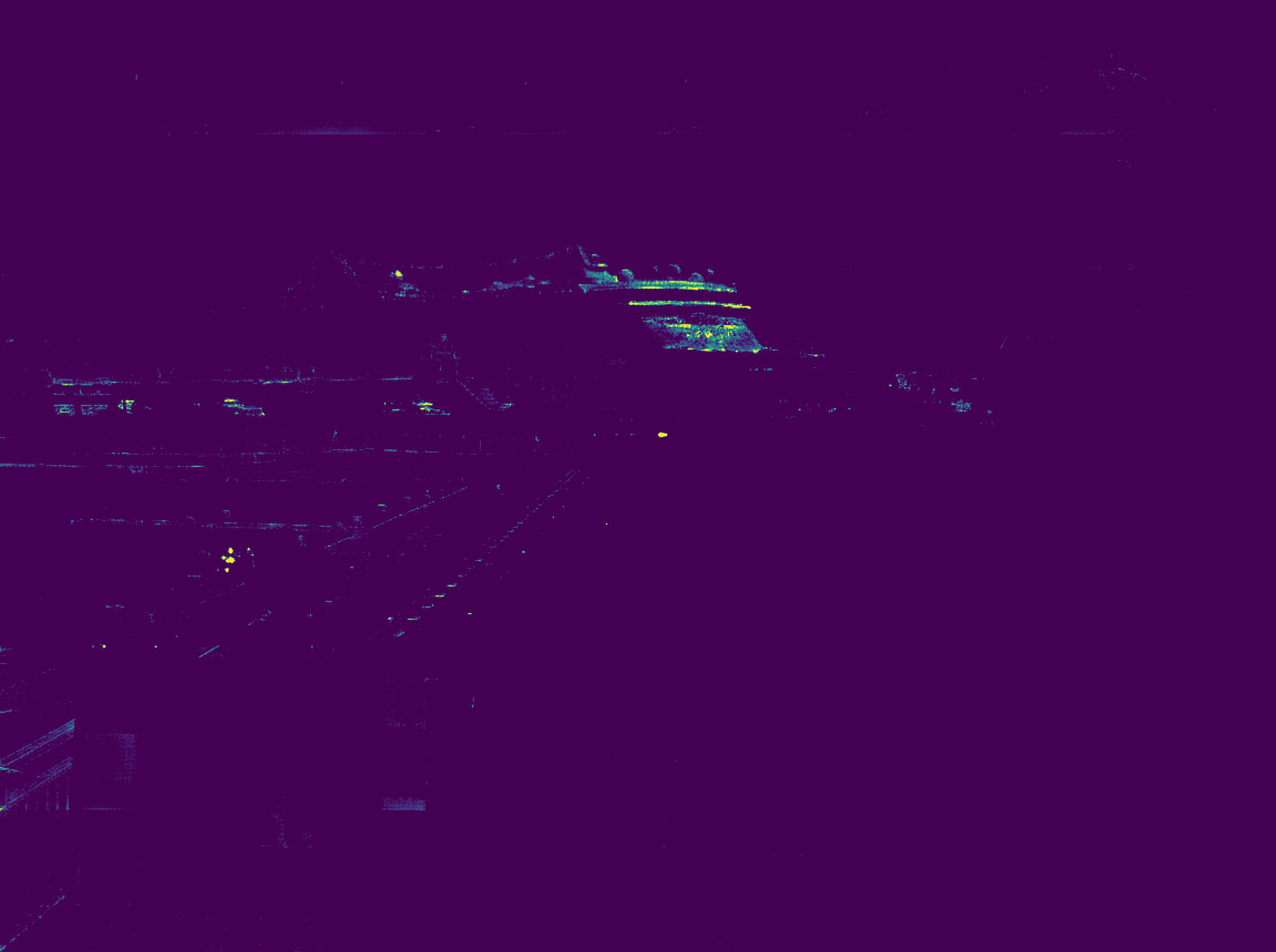}
            \includegraphics[width=0.03\textwidth,height=0.12\textwidth]{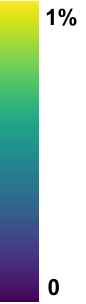}
            \includegraphics[width=0.18\textwidth,height=0.12\textwidth]{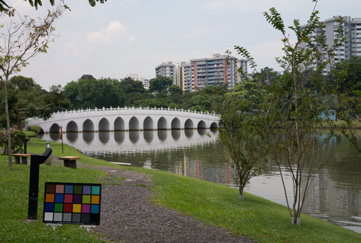}
            \includegraphics[width=0.18\textwidth,height=0.12\textwidth]{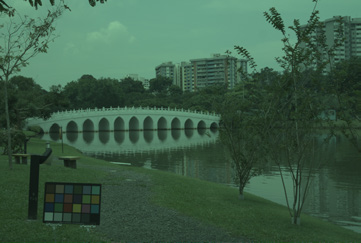}
            \includegraphics[width=0.18\textwidth,height=0.12\textwidth]{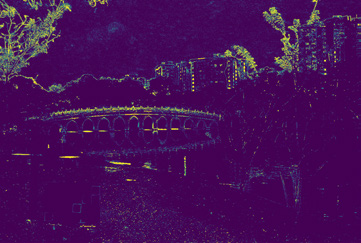}
            \includegraphics[width=0.18\textwidth,height=0.12\textwidth]{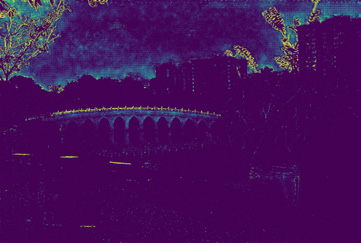}
            \includegraphics[width=0.18\textwidth,height=0.12\textwidth]{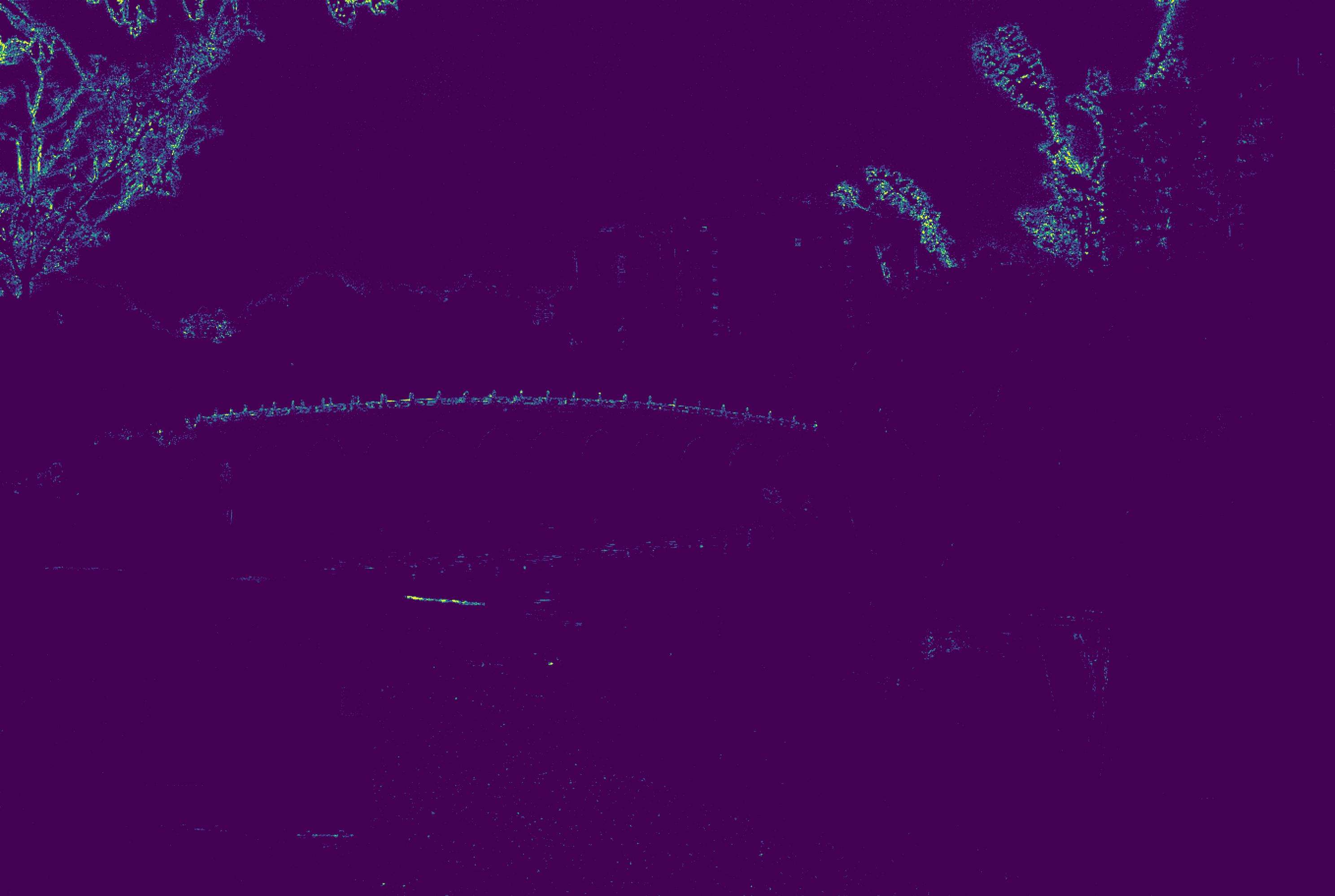}
            \includegraphics[width=0.03\textwidth,height=0.12\textwidth]{pics/img/colorbar.pdf}
            \includegraphics[width=0.18\textwidth,height=0.12\textwidth]{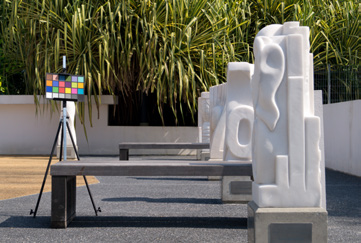}
            \includegraphics[width=0.18\textwidth,height=0.12\textwidth]{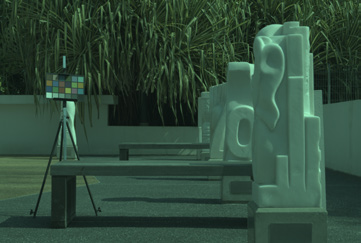}
            \includegraphics[width=0.18\textwidth,height=0.12\textwidth]{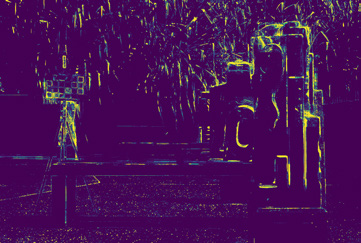}
            \includegraphics[width=0.18\textwidth,height=0.12\textwidth]{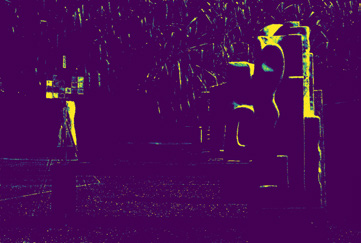}
            \includegraphics[width=0.18\textwidth,height=0.12\textwidth]{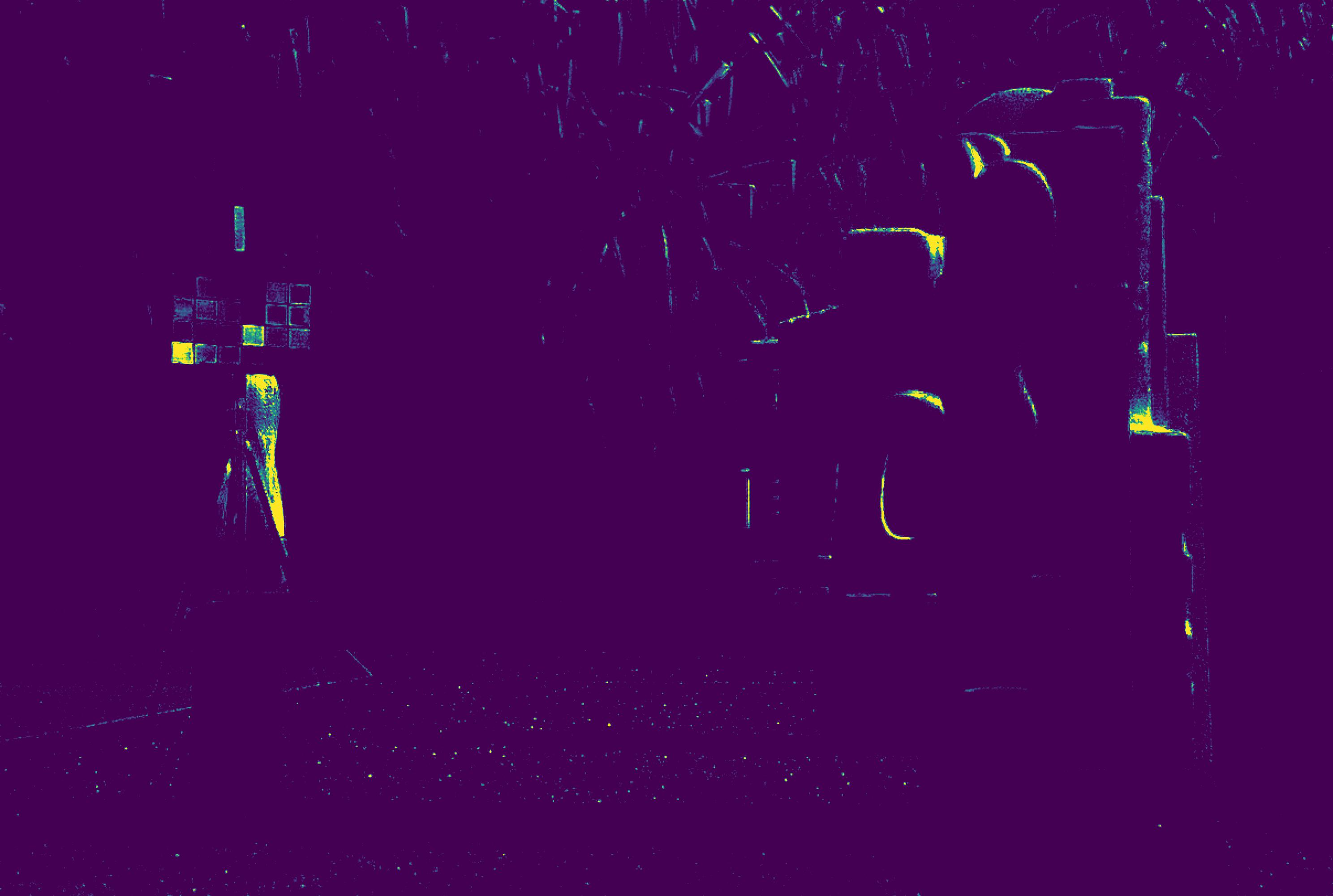}
            \includegraphics[width=0.03\textwidth,height=0.12\textwidth]{pics/img/colorbar.pdf}
            \parbox[c]{0.18\textwidth}{\centering sRGB}
            \parbox[c]{0.18\textwidth}{\centering GT}
            \parbox[c]{0.18\textwidth}{\centering SAM}
            \parbox[c]{0.18\textwidth}{\centering CAM}
            \parbox[c]{0.18\textwidth}{\centering Ours}
            \parbox[c]{0.03\textwidth}{\hspace{1mm}}
        \end{minipage}
        \label{fig:visual_image}
    } 
    \\ 
    \vspace{-1pt}
    \subfloat[Visual comparisons for \textbf{sRGB-to-RAW \textit{video} de-rendering}.]{%
        \begin{minipage}{\textwidth}
            \centering
            \footnotesize
            \includegraphics[width=0.18\textwidth,height=0.10\textwidth]{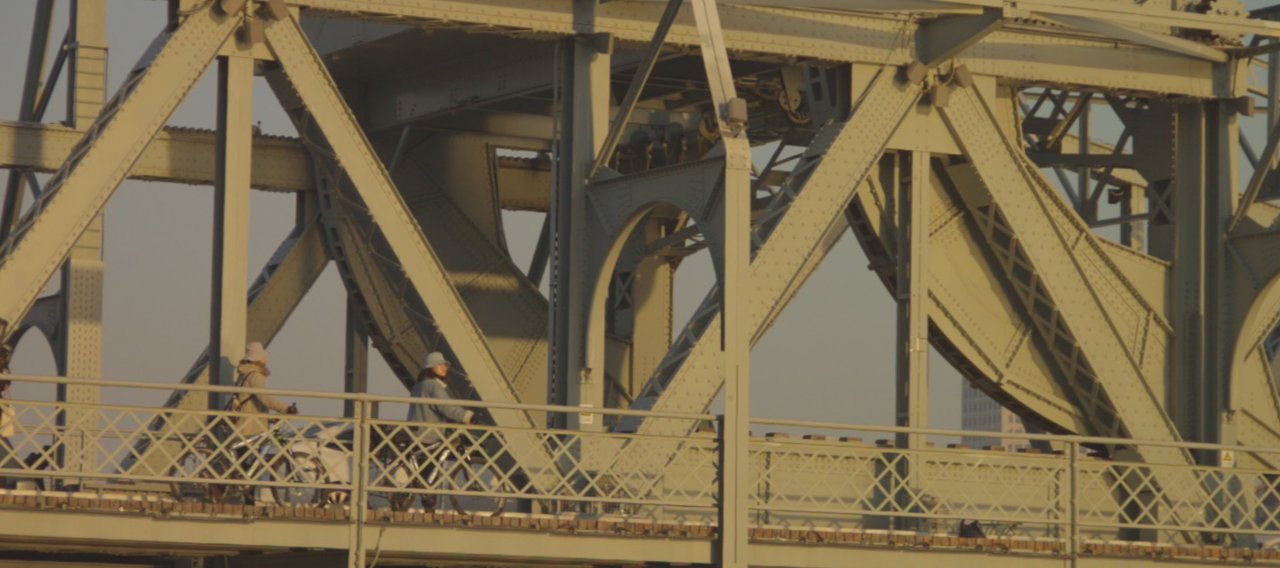}
            \includegraphics[width=0.18\textwidth,height=0.10\textwidth]{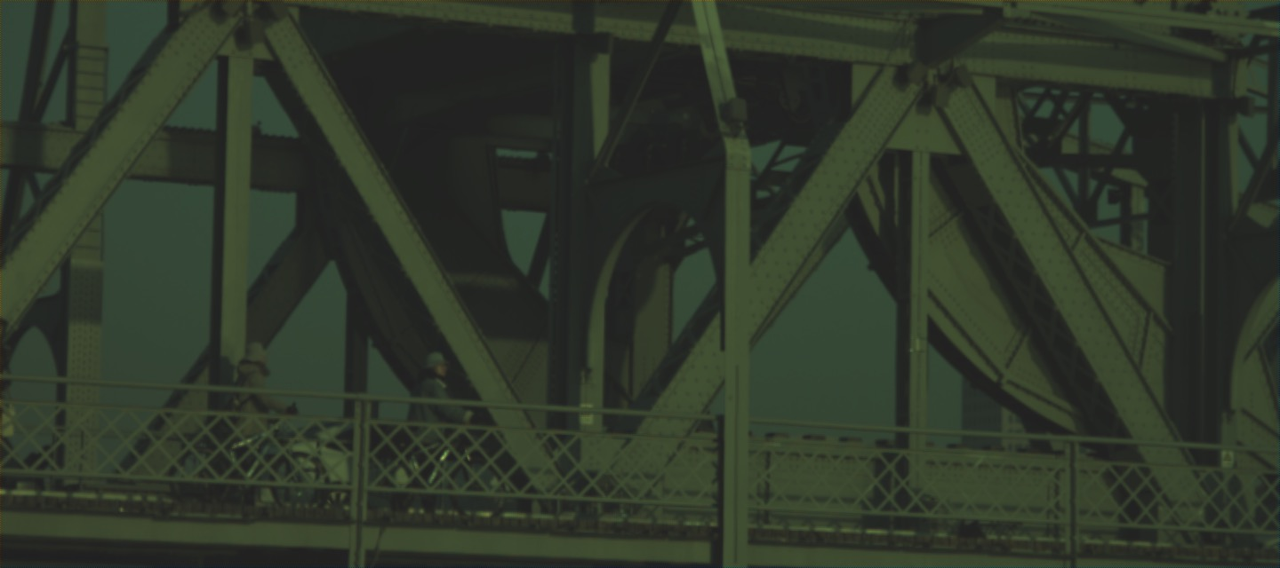}
            \includegraphics[width=0.18\textwidth,height=0.10\textwidth]{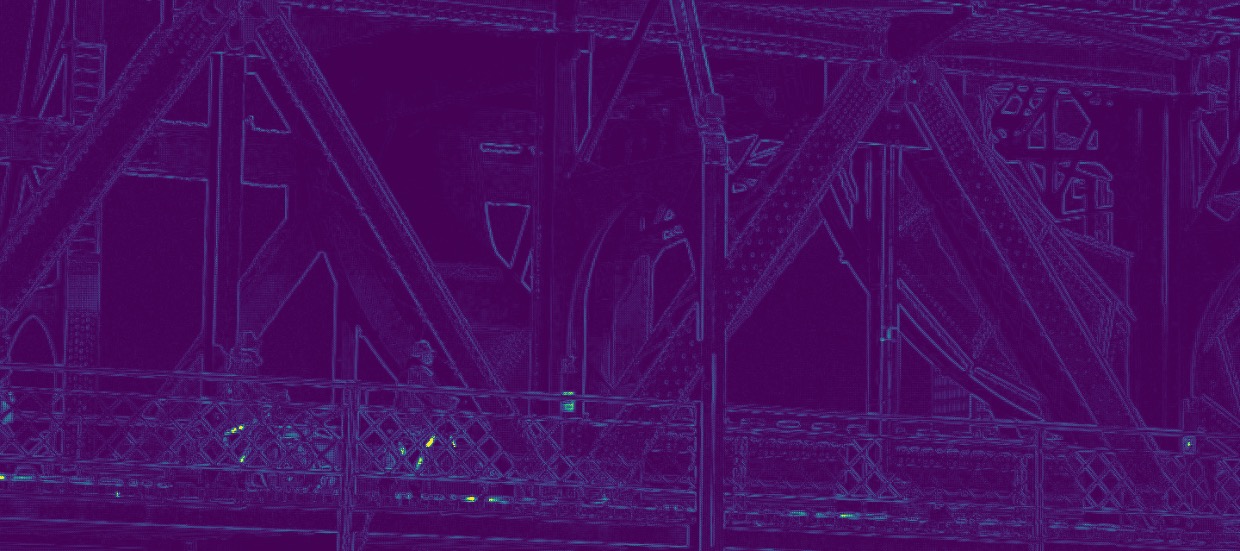}
            \includegraphics[width=0.18\textwidth,height=0.10\textwidth]{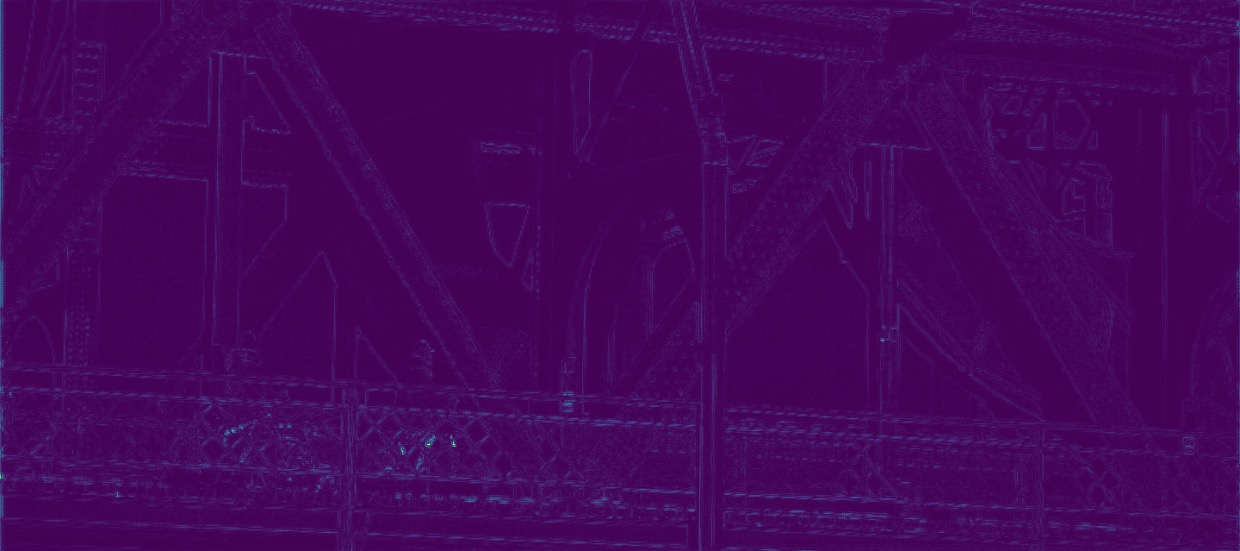}
            \includegraphics[width=0.18\textwidth,height=0.10\textwidth]{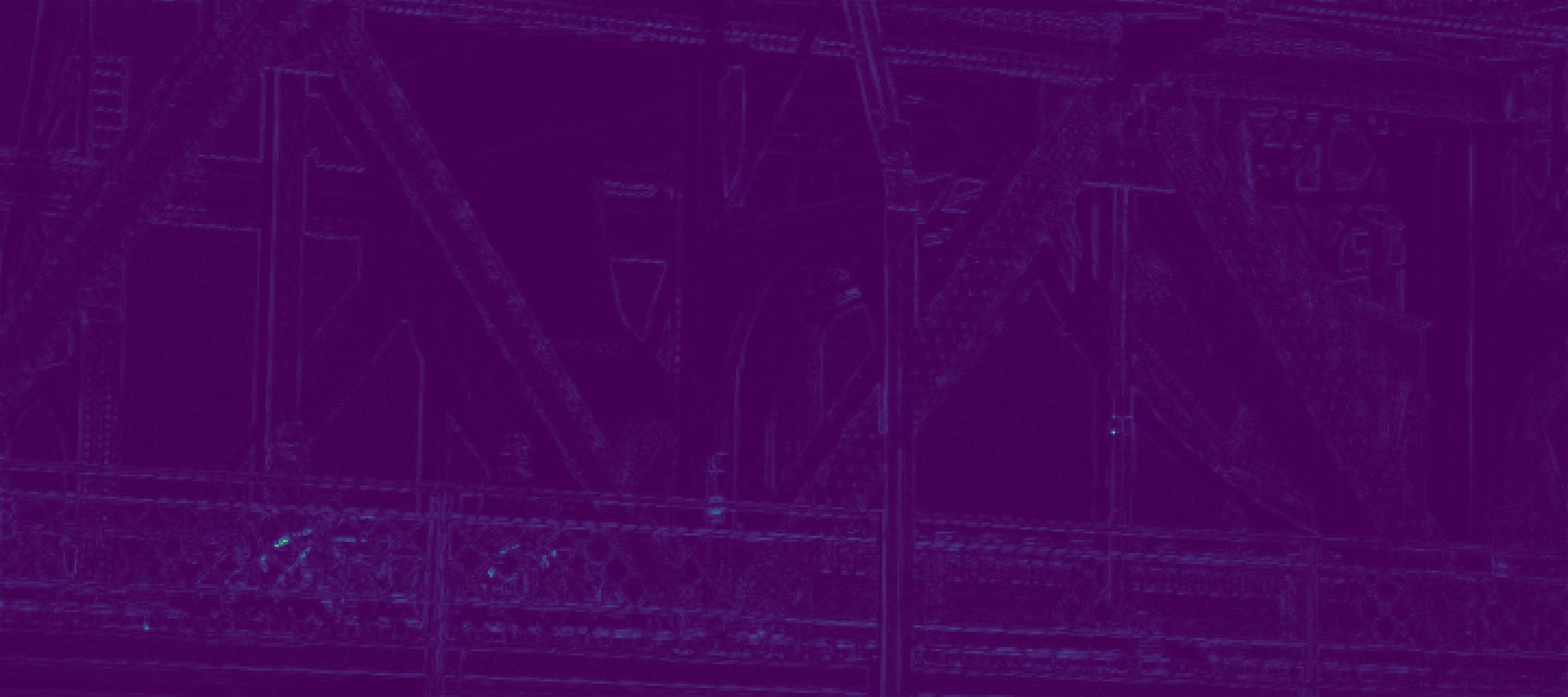}
            \includegraphics[width=0.03\textwidth,height=0.10\textwidth]{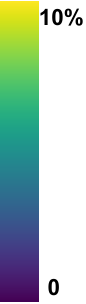}
            \includegraphics[width=0.18\textwidth,height=0.10\textwidth]{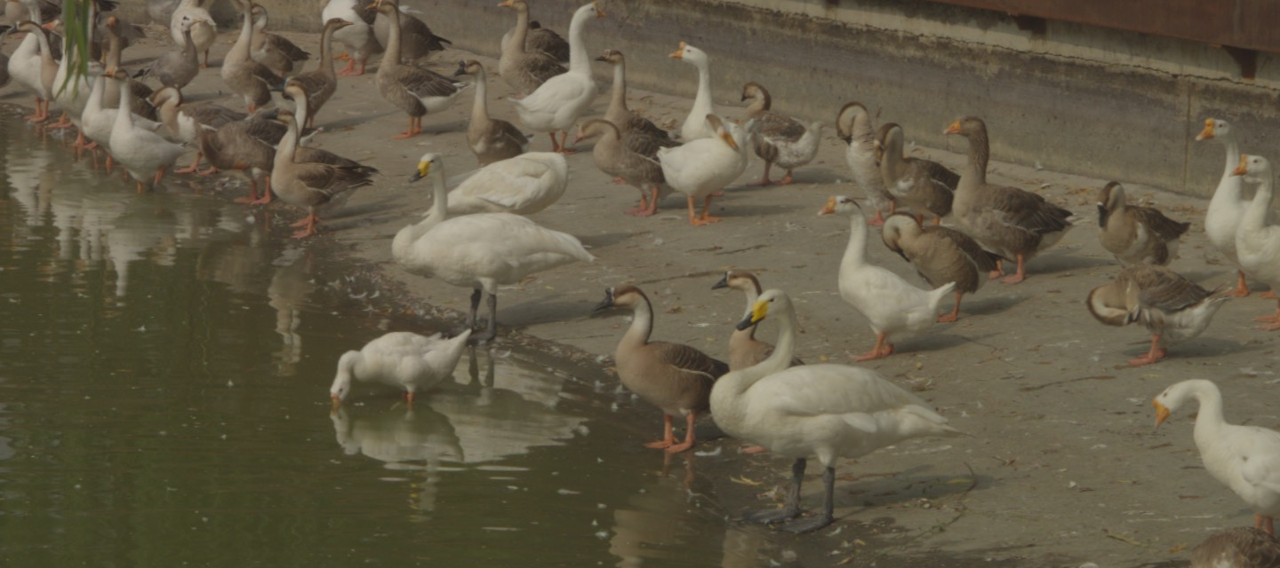}
            \includegraphics[width=0.18\textwidth,height=0.10\textwidth]{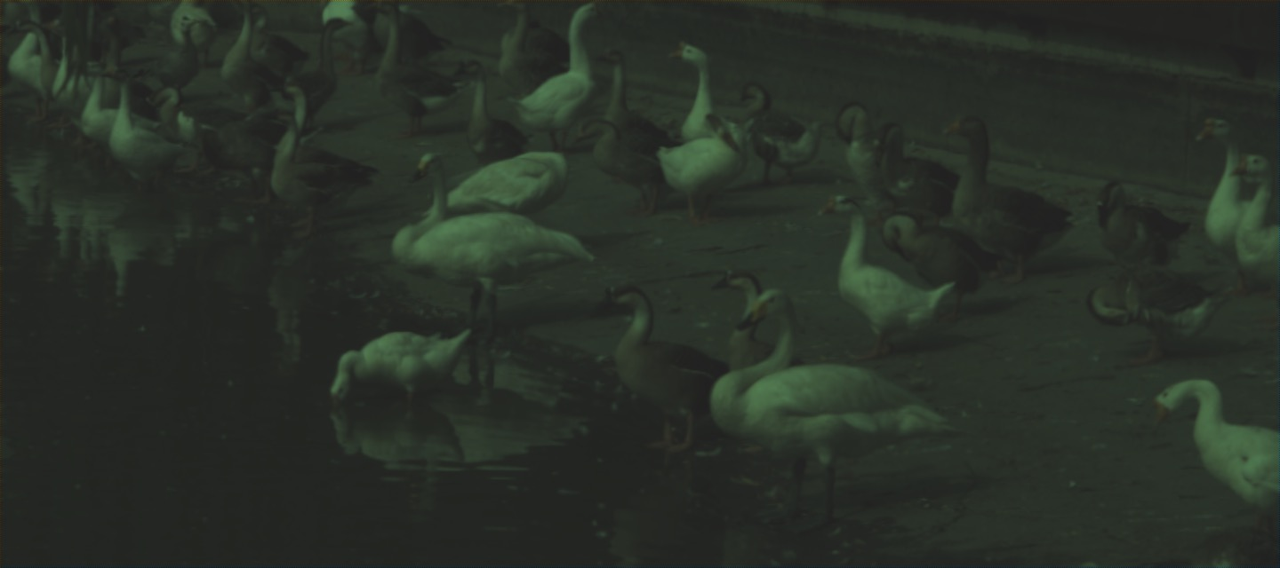}
            \includegraphics[width=0.18\textwidth,height=0.10\textwidth]{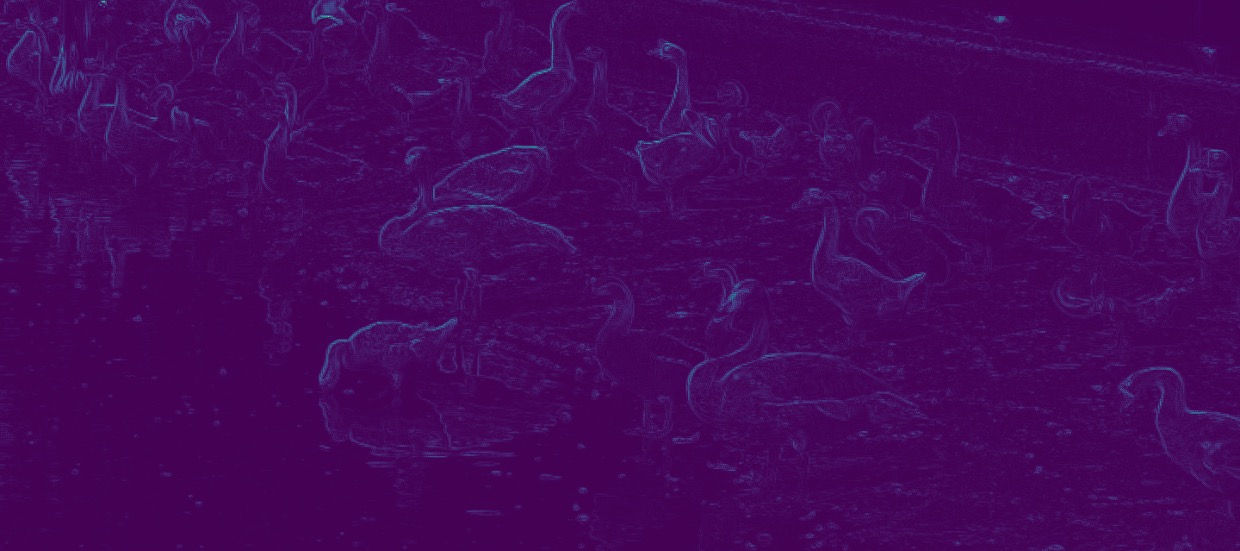}
            \includegraphics[width=0.18\textwidth,height=0.10\textwidth]{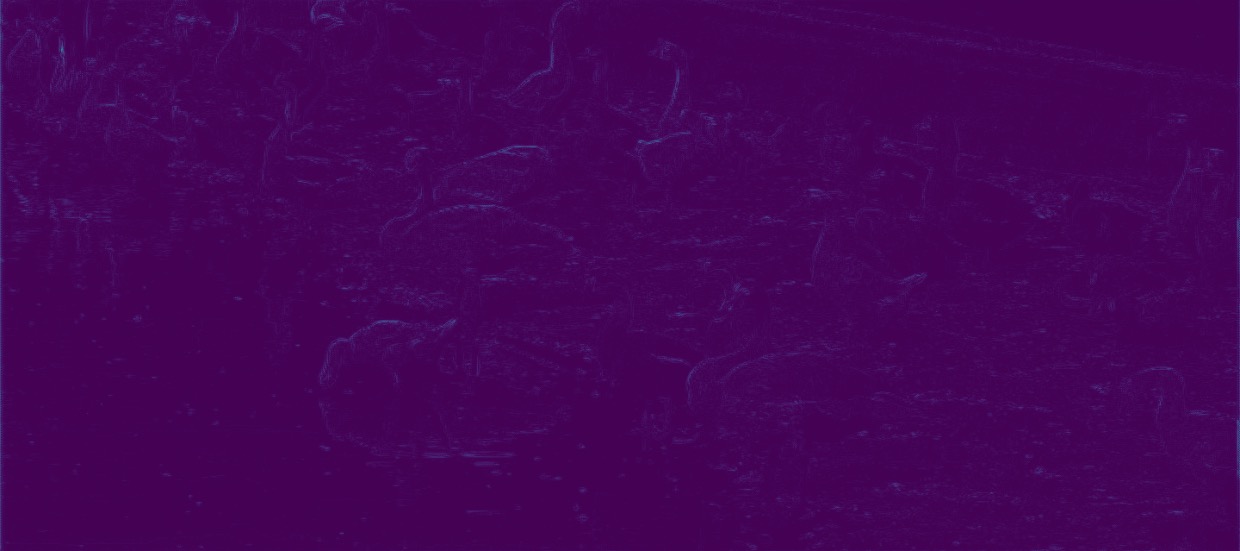}
            \includegraphics[width=0.18\textwidth,height=0.10\textwidth]{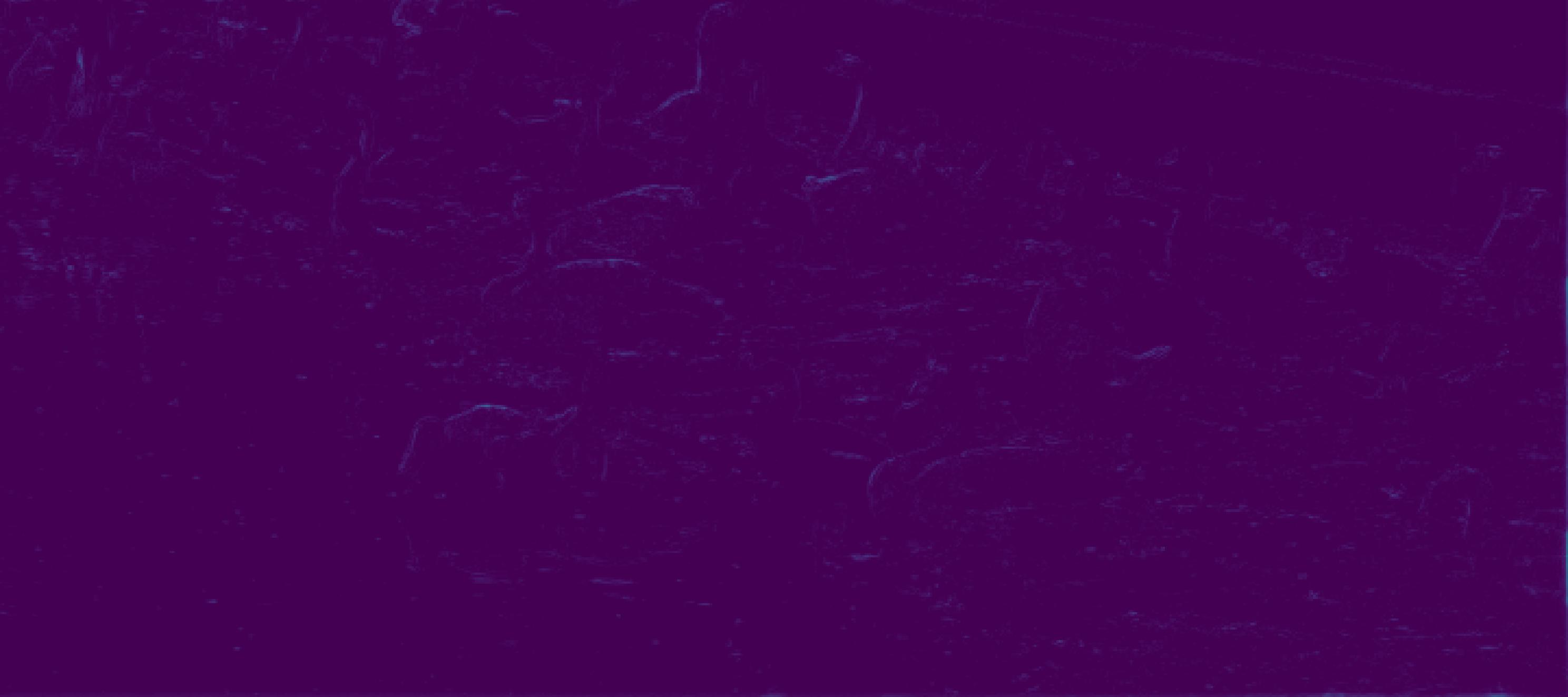}
            \includegraphics[width=0.03\textwidth,height=0.10\textwidth]{pics/video/colorbar_video.pdf}
            \parbox[c]{0.18\textwidth}{\centering sRGB}
            \parbox[c]{0.18\textwidth}{\centering GT}
            \parbox[c]{0.18\textwidth}{\centering INF} 
            \parbox[c]{0.18\textwidth}{\centering RVD}
            \parbox[c]{0.18\textwidth}{\centering Ours}
            \parbox[c]{0.03\textwidth}{\hspace{1mm}}
        \end{minipage}
        \label{fig:visual_video}
    } \vspace{-3mm}
	\caption{\textbf{Visual comparisons on image dataset CAM and video dataset RVD-part2.} As shown in this figure, our unified model achieves the highest accuracy and lowest error compared to other methods, across both image and video datasets.}
	\label{fig:vis}
    \vspace{-5mm}
\end{figure*}

\section{Experiments}

\subsection{Experiments setup}
\vspace{-1mm}
\textbf{Datasets.}
We conduct video-based experiments on the RVD dataset~\cite{videoraw}, specifically utilizing its currently available subset, RVD-Part2. This subset is collected data from \cite{realrawvsr}. We preserves the original splits of training and test sets. The training set comprises 130 videos, each containing approximately 50 frames, totaling 6,308 images, while the test set includes 20 videos with 983 images in total. All images have a resolution of 1440 × 640. The dataset uses original RAW data that has not undergone any ISP operations, and organizes the single-channel RAW data into a four-channel format: [R, Gr, Gb, B].
For image-based experiments, we use the same data settings as CAM, hereafter referred to as the CAM dataset. This dataset is captured from NUS dataset~\cite{cheng2014illuminant}, including images from three cameras (Samsung NX2000, Olympus E-PL6, and Sony SLT-A57), with 202, 208, and 268 raw images, respectively. Each raw Bayer image is demosaiced using standard bilinear interpolation to create a 3-channel raw-RGB image, which is then processed through a software ISP emulator~\cite{karaimer2016software} to generate the corresponding sRGB image. We follow the same training, validation, and test splits as CAM and crop all images into overlapping 128 × 128 patches for training.

\noindent
\textbf{Implements details.}
Our network is implemented on the PyTorch platform and trained on NVIDIA RTX 4090 GPUs. We use the Adam optimizer~\cite{kingma2014adam} with an initial learning rate of $2\times10^{-4}$, which is reduced by a factor of 0.1 every 20 epochs. In this study, we use the ConvNeXt~\cite{liu2022convnet} backbone pre-trained on ImageNet~\cite{deng2009imagenet} as the encoder, and the number of LTA-Mamba is set to 3. 
In image-based tasks, we train with a batch size of 64 for 80 epochs, applying a metadata sampling rate of 1.5\% as in CAM. In video-based tasks, input frames are randomly cropped to a spatial resolution of 256×256, with 5 frames per video clip, and training is conducted with a batch size of 4 for 50 epochs. 

\begin{table*}[t]
    \centering
    \setlength{\tabcolsep}{15pt}
    \resizebox{\linewidth}{!}{
    \begin{tabular}{c|cc|cc|cc|cc}
        \toprule
        \multirow{2}{*}{Method} & \multicolumn{2}{c|}{Samsung NX2000} & \multicolumn{2}{c|}{Olympus E-PL6} & \multicolumn{2}{c|}{Sony SLT-A57} & \multicolumn{2}{c}{Average} \\ \cline{2-3} \cline{4-5} \cline{6-7} \cline{8-9}
        & \multicolumn{1}{c}{PSNR $\uparrow$}   & SSIM $\uparrow$  & \multicolumn{1}{c}{PSNR $\uparrow$}  & SSIM $\uparrow$  & \multicolumn{1}{c}{PSNR $\uparrow$}  & SSIM $\uparrow$ & \multicolumn{1}{c}{PSNR $\uparrow$}  & SSIM $\uparrow$ \\
        \hline
        \hline
        RIR~\cite{nguyen2016raw}       & 45.66 & 0.9939 & 48.42 & 0.9924 & 51.26 & 0.9982 & 48.45 & 0.9948 \\
        SAM~\cite{spatiallyaware}      & 47.03 & 0.9962 & 49.35 & 0.9978 & 50.44 & 0.9982 & 48.94 & 0.9974 \\
        CAM~\cite{cam}                 & 48.08 & 0.9968 & 50.71 & 0.9975 & 50.49 & 0.9973 & 49.76 & 0.9972 \\
        CAM~\cite{cam} w/ fine-tuning  & \underline{49.57} & \underline{0.9975} & \underline{51.54} & \underline{0.9980} & \underline{53.11} & \underline{0.9985} & \underline{51.41} & \underline{0.9980} \\
        \rowcolor{gray!15}
        Ours & \textbf{51.04} & \textbf{0.9978} & \textbf{53.13} & \textbf{0.9984} & \textbf{55.22} & \textbf{0.9989} & \textbf{53.13} & \textbf{0.9984}\\
        \hline
        \bottomrule
    \end{tabular}
    }
    \vspace{-3mm}
    \caption{The quantitative results of CAM dataset~\cite{cam} conditioned on sRGB images. The $\uparrow$ indicates that larger values are preferable, with the best results highlighted in \textbf{bold}. \underline{Underlined} values represent the second-best results.}
    \vspace{-4mm}
    \label{tab:imaga_cam}
\end{table*}

\begin{table}[t]
    \centering
    \setlength\tabcolsep{6.3pt}
    \scalebox{0.92}{
    \begin{tabular}{c|c|cc}
        \toprule
        Method & Metadata & PSNR $\uparrow$  & SSIM $\uparrow$ \\
        \hline
        \hline
        CAM~\cite{cam}                & image &   42.25          & 0.9856 \\           
        CAM w/ fine-tuning~\cite{cam} & image &   42.26          & 0.9858 \\                 
        INF~\cite{li2023metadata}     & image &   46.25          & 0.9939 \\                  
        RVD~\cite{videoraw}           & video &   \underline{49.71}       & \textbf{0.9983} \\        
        \rowcolor{gray!15}
        Ours                          & video &   \textbf{51.97}          & \underline{0.9965} \\
        \hline
        \bottomrule
    \end{tabular}
    }
    \vspace{-3mm}
    \caption{The quantitative results of RVD-Part2 dataset~\cite{videoraw}. }
    \vspace{-4mm}
    \label{tab:image_rvd}
\end{table}

\subsection{Comparison with Image De-rendering Methods}

The comparison for sRGB-to-RAW image de-rendering is shown in \cref{tab:imaga_cam}.
It is important to note that neither RIR nor SAM has made their source code publicly available. Therefore, we rely on the reproduced results provided by CAM~\cite{cam} for comparison. In the table, the third row shows the performance of CAM's offline model, while the fourth row displays the performance of CAM's model after additional online fine-tuning during inference. Since the online fine-tuning approach does not align with the goal of minimizing the complexity of model deployment, our method does not incorporate any online learning steps. Using the same sampling model configuration as CAM, our approach significantly outperforms both the original CAM model and its online fine-tuned version across all three datasets. Specifically, under offline training conditions, we achieve a PSNR improvement of 2.96 dB on the Samsung NX2000 camera, 2.42 dB on the Olympus E-PL6, and 4.73 dB on the Sony SLT-A57. The superior reconstruction quality is illustrated in \cref{fig:visual_image}, highlighting the enhanced capability of our unified model in sRGB-to-RAW de-rendering compared to previous task-specific methods.

\subsection{Comparison with Video De-rendering Methods}
We further evaluate the effectiveness of RAWMamba for video sRGB-to-RAW de-rendering. \cref{tab:image_rvd} presents the numerical performance on the RVD-Part2 dataset, while \cref{fig:visual_video} illustrates the visual comparison. The video sRGB-to-RAW de-rendering task was newly introduced in the RVD~\cite{videoraw}, and we directly compare our results with those reported in the paper. In their experiments, the CAM~\cite{cam} and INF~\cite{li2023metadata} methods were retrained and tested on the RVD-Part2 dataset. As shown in the comparison, image-specific methods such as CAM and INF significantly underperform when contrasted with RVD, which was specifically designed for video tasks. Compared to the state-of-the-art method RVD, our approach achieves a 2.26 dB improvement in PSNR, demonstrating that our method not only excels in efficiently handling image-based data but also exhibits robust performance in video data.

\begin{table}[t]
    \centering
    \setlength\tabcolsep{8pt}
    \scalebox{0.945}{
    \begin{tabular}{cc|cc|cc}
        \toprule
        \multicolumn{2}{c|}{UME} & \multicolumn{2}{c|}{LTA-Mamba} & \multirow{2}{*}{PSNR $\uparrow$}  & \multirow{2}{*}{SSIM $\uparrow$} \\
        \cline{1-4}
        GEB & LEB & Global & Local  & &  \\
        \hline
        \hline
        & & &              & 48.83 & 0.9956 \\
        \checkmark & & &              & 51.91 & 0.9982 \\                    
        & \checkmark & &   & 52.03 & 0.9982 \\                    
        \checkmark & \checkmark & &   & 52.68 & 0.9983 \\   

        \hline

        \checkmark & \checkmark & \checkmark &   & 54.01 & 0.9987 \\        
        \checkmark & \checkmark & &  \checkmark  & 54.25 & 0.9987 \\         
        \rowcolor{gray!15}
        \checkmark & \checkmark & \checkmark & \checkmark  & \textbf{55.22} & \textbf{0.9989} \\
        \hline
        \bottomrule
    \end{tabular}
    }
    \vspace{-3mm}
    \caption{The ablation study of the introduced modules. Results were obtained on the Sony SLT-A57 camera of the CAM dataset.   }
    \vspace{-5mm}
    \label{tab:imaga_ablation}
\end{table}

\subsection{Ablation study}
\noindent
\textbf{Effectiveness of UME.} We first investigate the role of the UME. We conduct a series of experiments, shown in \cref{tab:imaga_ablation}, by removing GEB and LEB from the full pipeline. The first three rows in the table demonstrate the critical role of metadata in reconstruction, where GEB and LEB can effectively extract reference information from metadata. The fourth row further highlights that integrating both local and global perspectives from metadata enables more effective and comprehensive extraction of reference information.

\noindent
\textbf{Effectiveness of LTA-Mamba.}
We further study the design of the proposed LTA-Mamba by removing global scan part and local scan part in the LTA-Mamba. The fifth and sixth rows demonstrate that both the local and global scanning based Mamba blocks enhance model’s ability to express global consistency, thereby more effectively utilizing the reference information from sparse metadata. The final row confirms that the Mamba module, designed with local tone mapping, is better suited for sRGB-to-RAW de-rendering tasks compared to traditional Mamba structures.




\section{Conclusion}
In this paper, we propose a Mamba-based model, RAWMamba, for unified sRGB-to-RAW de-rendering across both image and video inputs. Unlike previous task-specific sRGB-to-RAW methods, RAWMamba introduces a Unified Metadata Embedding (UME) module that consolidates diverse metadata into a unified representation and leverages a multi-perspective affinity modeling approach to enhance metadata extraction. Furthermore, we incorporate the Local Tone-Aware Mamba (LTA-Mamba) module, designed to capture long-range dependencies and facilitate effective global metadata propagation. Experimental results demonstrate that RAWMamba achieves state-of-the-art performance on both video and image datasets, identifying the  superior effectiveness and robustness of our method.

{ \small
    \bibliographystyle{ieeenat_fullname}
    \bibliography{main}
}


\end{document}